\documentclass[10pt,twocolumn,letterpaper]{article}

\usepackage[pagenumbers]{iccv} 
\usepackage{colortbl} 
\usepackage{multirow}
\usepackage{makecell} 
\usepackage{tabularx}
\usepackage{graphicx} 
\usepackage{booktabs}

\usepackage{graphicx}  

\usepackage{graphicx}
\usepackage{multicol}
\usepackage{multirow}
\usepackage{adjustbox}
\usepackage{graphicx}
\usepackage{amsmath}
\usepackage[pagebackref,breaklinks,colorlinks,allcolors=iccvblue]{hyperref}
\usepackage[capitalize,noabbrev]{cleveref} 
\usepackage{booktabs}
\usepackage{subcaption}
\usepackage{colortbl}

\definecolor{lightergray}{gray}{0.9} 
\definecolor{blue}{RGB}{65, 105, 225}
\definecolor{orange}{RGB}{255, 229, 204}

\usepackage{url}           
\usepackage{bm}
\definecolor{iccvblue}{rgb}{0.21,0.49,0.74}

\def\paperID{} 
\def\confName{ICCV}
\def\confYear{2025}

\newcommand{\ours}{\texttt{QuantCache}}

\title{$\ours$: Adaptive Importance-Guided Quantization\\with Hierarchical Latent and Layer Caching for Video Generation}

\author{
  Junyi Wu$^{1}$\thanks{Equal contribution}~,\enspace
  Zhiteng Li$^{1}$\footnotemark[1]~,\enspace
  Zheng Hui$^{2}$,\enspace
  Yulun Zhang$^{1}$\footnotemark[2]~,\enspace
  Linghe Kong$^{1}$,\enspace 
  Xiaokang Yang$^{1}$~\enspace
  \\
  \textsuperscript{1}Shanghai Jiao Tong University,\enspace
  \textsuperscript{2}MGTV, Shanhai Academy\enspace\\
    \vspace{-5mm}
}

\begin{document}

\twocolumn[{%
\renewcommand\twocolumn[1][]{#1}%
\maketitle
\begin{center}
    \centering
    \captionsetup{type=figure}
    \vspace{-3mm}
    \includegraphics[width=0.9\linewidth]{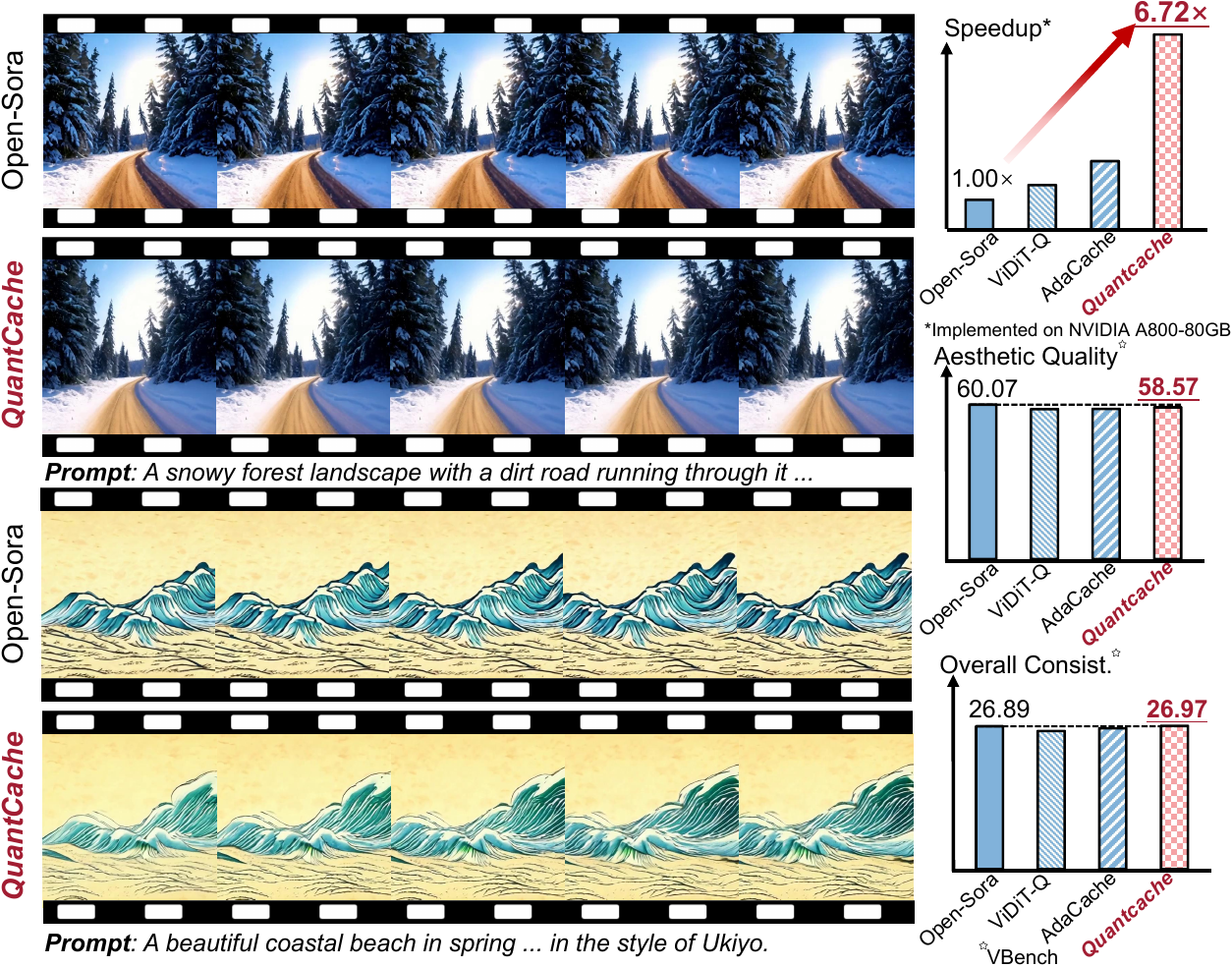}
    \vspace{-3mm}
   \caption{$\ours$ is a training-free acceleration framework with an end-to-end \textbf{6.72$\times$} speedup against Open-Sora~\cite{opensora}. Compared with ViDiT-Q~\cite{zhao2024viditq} and AdaCache~\cite{adacache}, $\ours$ achieves superior quality scores, demonstrating the effectiveness of our method.}
   
    \label{fig:teaser}
    
\end{center}%
}]

\let\thefootnote\relax
\footnotetext{*\,Equal contribution.}
\footnotetext{\dag\,Corresponding author: Yulun Zhang, yulun100@gmail.com}

\begin{abstract}
Recently, Diffusion Transformers (DiTs) have emerged as a dominant architecture in video generation, surpassing U-Net-based models in terms of performance. However, the enhanced capabilities of DiTs come with significant drawbacks, including increased computational and memory costs, which hinder their deployment on resource-constrained devices. Current acceleration techniques, such as quantization and cache mechanism, offer limited speedup and are often applied in isolation, failing to fully address the complexities of DiT architectures. In this paper, we propose $\ours$, a novel training-free inference acceleration framework that jointly optimizes hierarchical latent caching, adaptive importance-guided quantization, and structural redundancy-aware pruning. $\ours$ achieves an end-to-end latency speedup of \textbf{6.72$\times$} on Open-Sora with minimal loss in generation quality. Extensive experiments across multiple video generation benchmarks demonstrate the effectiveness of our method, setting a new standard for efficient DiT inference. The code and models will be available at \url{https://github.com/JunyiWuCode/QuantCache}.

\end{abstract}

\setlength{\abovedisplayskip}{2pt}
\setlength{\belowdisplayskip}{2pt}

\vspace{-2.5mm}
\section{Introduction}
\label{sec:intro}
\vspace{-1.5mm}

Recently, Diffusion Transformers (DiTs)~\cite{Peebles_2023_ICCV,yang2024cogvideox} have gained significant attention due to their superior performance in generative modeling, particularly in video generation tasks~\cite{NEURIPS2020_4c5bcfec,pmlr-v139-nichol21a,kingma2021on,Karras2022ElucidatingTD,Song2021DenoisingDI,chen2022analog,Salimans2022ProgressiveDF}. These models leverage the powerful attention mechanism of transformers, which allows them to capture long-range dependencies and produce high-quality outputs. However, this remarkable performance comes at the cost of substantial computational and memory requirements, which hinder their practical deployment, especially on resource-constrained devices. For instance, generating a 64-frame, 512$\times$512 resolution video with the Open-Sora model~\cite{opensora} on an NVIDIA A800-80GB GPU takes up to 130 seconds. The computational complexity is compounded by the quadratic growth of attention mechanisms in DiTs, which increases with the fixed long timesteps. Therefore, despite their impressive generative capabilities, DiTs face significant barriers to efficient deployment in real-world and edge device applications.

Model quantization is a widely used technique for reducing the memory and computational overhead of large models by compressing weights and activations into low-bit representations. Among the various quantization approaches, Post-Training Quantization (PTQ)~\cite{zhao2024viditq,chen2024QDiT,li2024svdquant,dong2024ditas} is particularly appealing due to its minimal training requirements and rapid deployment capabilities, making it ideal for large models like DiTs. Unlike Quantization-Aware Training (QAT)~\cite{lu2024terdit}, which requires extensive fine-tuning resources, PTQ allows for efficient quantization with much fewer computational costs. In addition to quantization, inference acceleration frameworks have been explored to further optimize diffusion models, including distillation~\cite{li2024hunyuandit}, pruning~\cite{fang2024tinyfusion}, and cache-based methods~\cite{chen2024deltadittrainingfreeaccelerationmethod, adacache}.

However, these methods are often applied in isolation and neglect to fully leverage the synergies among them. A major limitation of existing approaches is their reliance on static heuristics, which do not adapt to the dynamic nature of the diffusion process. For instance, uniform quantization strategies apply a fixed bit-width across all layers and timesteps. They ignore that different layers exhibit varying levels of importance depending on the generation stage. Similarly, existing caching strategies use predefined schedules, neglecting to account for frame-specific variations in content evolution. These issues motivate the need for a more adaptive approach that can dynamically allocate computational resources based on specific content analyses.

To address those challenges, we propose a joint optimization framework $\ours$ for video generation. First, we propose hierarchical latent caching (HLC) that adaptively determines when to refresh cached features based on inter-step feature divergence. HLC reduces redundant computations while preserving generation quality. Second, we propose adaptive importance-guided quantization (AIGQ), where bit-widths are adjusted per timestep and per layer according to feature sensitivity. AIGQ ensures that more critical computations retain higher bit-widths while redundant ones are processed at lower bit-widths. Finally, we propose structural redundancy-aware pruning (SRAP) that selectively prunes layers with highly correlated feature representations within the same timestep, further reducing computational cost. By jointly optimizing these three techniques, $\ours$ effectively minimizes redundant computations while preserving the expressiveness of DiTs. Our contributions can be summarized as follows:
\begin{itemize}
    \item We propose an efficient video generation framework $\ours$ by jointly optimizing caching, quantization, and pruning. $\ours$ achieves \textbf{6.72$\times$} speedup against Open-Sora~\cite{opensora}~(\cref{fig:teaser}), surpassing SOTA methods while maintaining high generation quality.
    \item We propose hierarchical latent caching (HLC) that dynamically adjusts caching schedules based on feature divergence. HLC significantly reduces redundant computations in Diffusion Transformers (DiTs).
    \item We propose adaptive importance-guided quantization (AIGQ), a novel adaptive quantization framework that allocates precision levels based on timestep significance.
    \item We propose structural redundancy-aware pruning (SRAP), an online layer pruning method that selectively omits redundant computations within each timestep.
\end{itemize}

\vspace{-2mm}
\section{Related Works}
\vspace{-1.5mm}

\subsection{Diffusion Transformers}
\vspace{-1.5mm}

Diffusion Transformers (DiTs)~\cite{NEURIPS2020_4c5bcfec,pmlr-v139-nichol21a,kingma2021on,Karras2022ElucidatingTD,Song2021DenoisingDI,chen2022analog,Salimans2022ProgressiveDF} have emerged as a compelling alternative to traditional U-Net architectures in generative modeling tasks.
DiTs utilize the self-attention mechanism~\cite{attention} to effectively capture long-range dependencies, thereby enhancing the quality of generated visual content. For instance, Open-Sora~\cite{opensora} integrates a Variational Autoencoder (VAE)~\cite{kingma2022autoencodingvariationalbayes} with DiTs, enabling efficient high-quality video generation. Despite their notable success, transformer-based diffusion models face challenges related to computational complexity and memory consumption. The self-attention mechanism's computational requirements scale quadratically with the input size, making high-resolution image and multi-frame video generation particularly resource-intensive. Addressing these challenges is crucial for the practical deployment of such models, especially with limited computational resources.

\vspace{-1.5mm}
\subsection{Model Quantization}
\vspace{-1.5mm}
Model quantization~\cite{deng2024vq4ditefficientposttrainingvector,xiao2023smoothquant,li2024svdquant} is a pivotal technique for enhancing the efficiency of deep learning models by converting full-precision weights and activations into lower-bit representations, thereby reducing both memory footprint and computational load. Post-Training Quantization (PTQ)~\cite{li2024arb} stands out as a particularly effective approach, enabling the compression of pre-trained models without necessitating extensive retraining. In the realm of diffusion models, PTQ has been successfully applied to U-Net-based architectures~\cite{Shang_2023_CVPR}. The advent of Diffusion Transformers (DiTs) has further propelled advancements in generative modeling, offering superior scalability and performance. However, the application of PTQ to DiTs presents unique challenges due to their architectural distinctions from U-Net-based models. Addressing this,  ViDiT-Q~\cite{zhao2024viditq} proposed a quantization scheme tailored for DiTs that achieves lossless 8-bit weight and activation quantization, resulting in significant memory and latency improvements. These advancements underscore the critical role of specialized PTQ methods in optimizing the performance and efficiency of diffusion models, particularly as architectures evolve from U-Net-based structures to transformer-based designs.
\vspace{-1.5mm}
\subsection{Cache Mechanism}
\vspace{-1.5mm}
In the realm of diffusion models, cache-based acceleration techniques have been developed to enhance inference efficiency by reusing computations across timesteps. For instance, DeepCache~\cite{Ma_2024_CVPR} leverages temporal redundancy in U-Net architectures by caching high-level features during the denoising process, achieving significant speedups without necessitating model retraining.  For DiT architectures, AdaCache~\cite{adacache} introduces a training-free method tailored for video Diffusion Transformers (DiTs), implementing a content-dependent caching schedule that adapts to each video's complexity, thereby optimizing the quality-latency trade-off. $\Delta$-DiT~\cite{chen2024deltadittrainingfreeaccelerationmethod} introduces a specialized caching mechanism known as $\Delta$-Cache, designed specifically for DiT architectures. This approach involves analyzing the role of each DiT block in image generation and selectively reuses feature offsets, accelerating inference without compromising image quality. However, optimizing these cache strategies remains challenging, particularly in balancing efficiency gains with the preservation of generation quality.


\begin{figure*}[t]
    \centering
    \includegraphics[width=1\linewidth]{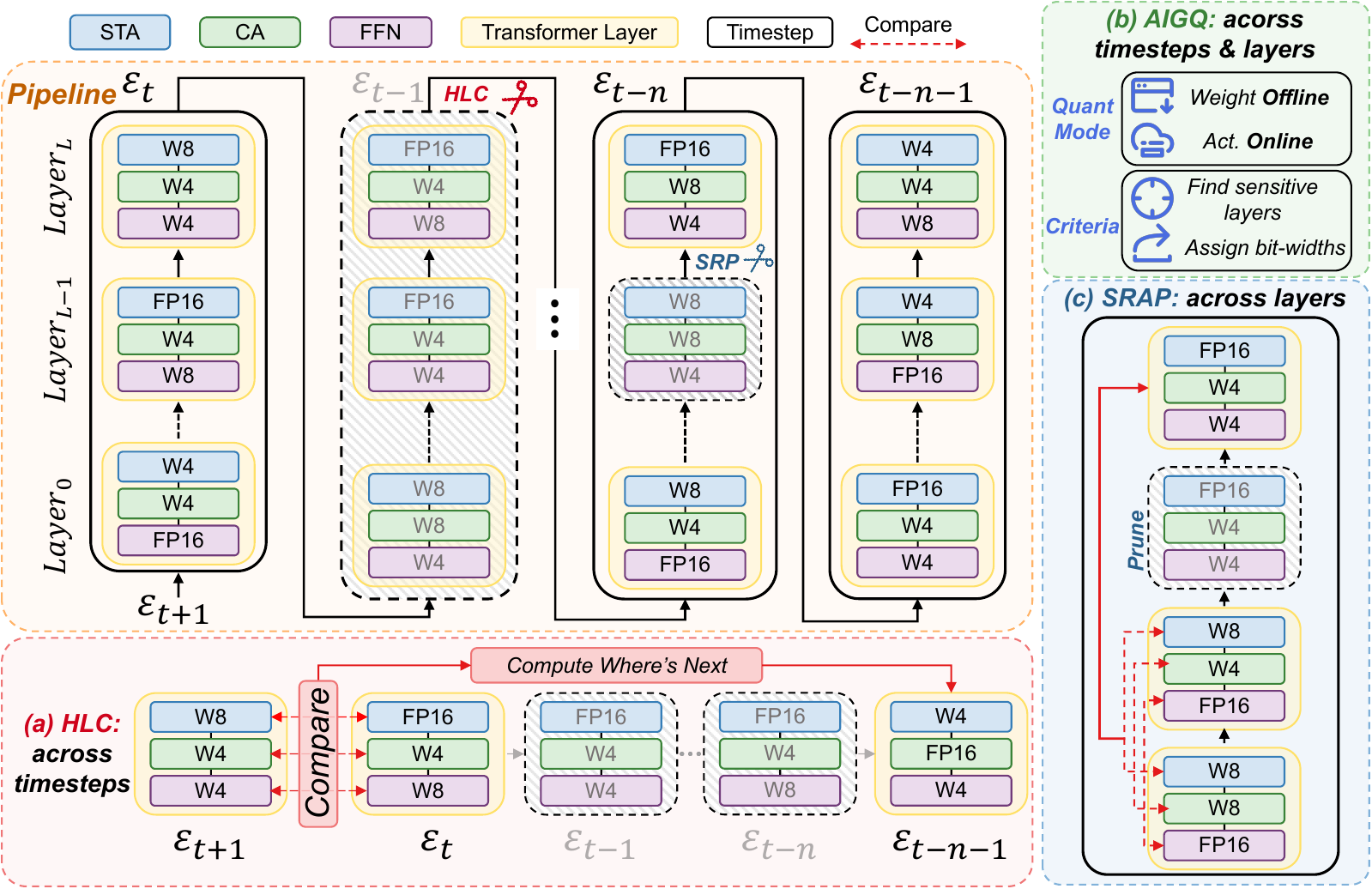}
    \vspace{-7mm}
    \caption{The overview of $\ours$ with (a) HLC, (b) AIGQ, (c) SRAP. STA, CA, and FFN respectively refer to spatial-temporal attention, cross attention, and feedforward network in a Transformer layer.    
 }
    \label{fig:overview}
    \vspace{-5.5mm}
\end{figure*}

\vspace{-1.5mm}
\section{Method}
\vspace{-1.5mm}
\subsection{Preliminary}
\vspace{-1.5mm}
\noindent\textbf{Diffusion Models.$\quad$}
Diffusion models~\cite{NEURIPS2020_4c5bcfec,pmlr-v139-nichol21a,kingma2021on,Karras2022ElucidatingTD,Song2021DenoisingDI,chen2022analog,Salimans2022ProgressiveDF} are a class of generative models inspired by the process of diffusion in physics, where particles spread out over time due to random motion. In the context of generative modeling, diffusion models operate by gradually adding noise to data in a forward process, and then reversing this process to reconstruct the original data.

The forward diffusion process starts with a clean data sample \( x_0 \sim q(x) \) and progressively adds noise over \( T \) timesteps. The noisy data at timestep \( t \) is defined as:
\begin{equation}
    x_t = \sqrt{\bar{\alpha}_t} x_0 + \sqrt{1 - \bar{\alpha}_t} \epsilon_t, \quad \epsilon_t \sim \mathcal{N}(0, I),
\end{equation}
where \( \bar{\alpha}_t \) is a schedule controlling the noise level at timestep \( t \), and \( \epsilon_t \) is Gaussian noise. As the diffusion process progresses, \( x_t \) becomes increasingly noisy, and by the final timestep \( T \), it is pure Gaussian noise.

The reverse diffusion process aims to recover the clean data \( x_0 \) from the noisy data \( x_t \). The model learns a parameterized distribution \( p_\theta(x_{t-1} | x_t) \), which predicts the clean data at timestep \( t-1 \) based on the noisy data at timestep \( t \). This is typically modeled as a Gaussian distribution:
\begin{equation}
    p_\theta(x_{t-1} | x_t) = \mathcal{N}\left(x_{t-1}; \mu_\theta(x_t, t), \Sigma_\theta(x_t, t)\right),
\end{equation}
where \( \mu_\theta(x_t, t) \) and \( \Sigma_\theta(x_t, t) \) are the mean and covariance predicted by the model at timestep \( t \).

In a typical Diffusion Transformers~\cite{yang2024cogvideox,Peebles_2023_ICCV,flux2024,hatamizadeh2024diffit,lin2022genie,kim2024versatile}, the noisy data \( x_t \) is processed through a sequence of Transformer blocks. Each block consists of self-attention (SA), cross-attention (CA), and feed-forward networks (FFN). The self-attention mechanism allows the model to capture complex dependencies within the data, while the cross-attention layers integrate additional conditional information, like class labels or textual descriptions.

\noindent\textbf{Model Quantization.$\quad$}
Quantization~\cite{DBLP:journals/corr/abs-2106-08295} is a technique employed to reduce the computational and memory demands by representing weights and activations with lower precision. This process involves approximating high-precision values with low-bit representations, thereby accelerating inference and decreasing storage requirements.

Formally, consider a neural network with $L$ layers, where each layer $l$ has weights $W^{(l)}$ and activations $X^{(l)}$. The objective of quantization is to find proper bit-widths
that minimizes the discrepancy between the original and quantized models. 
In uniform quantization, both weights and activations are mapped to discrete levels within a fixed range. 
The quantization function $Q$ for a tensor $x$ with $b$-bit representation is defined as:
\begin{equation}
x_{\text{int}} = Q(x; s, z, b) = \text{clamp}\left( \left\lfloor \frac{x}{s} \right\rceil + z, 0, 2^b - 1 \right),
\end{equation}
where $s$ is the scaling factor, $z$ is the zero point, $\left\lfloor \cdot \right\rceil$ denotes rounding to the nearest integer, and $\text{clamp}(\cdot, a, c)$ restricts the values to the interval $[a, c]$. The scaling factor $s$ is typically determined by the range of $x$:
\begin{equation}
s = \frac{\max(x) - \min(x)}{2^b - 1}.
\end{equation}

Quantization error arises from two primary sources: clipping (or clamping) error and rounding error. Clipping error occurs when the dynamic range of $x$ exceeds the representable range, leading to saturation. Rounding error results from mapping continuous values to discrete levels. 
These errors are influenced by factors such as the bit-width $b$, the distribution of $x$, and the chosen quantization parameters.

In Diffusion Transformers (DiTs), quantization presents unique challenges. The isotropic architecture of DiTs~\cite{chen2024deltadittrainingfreeaccelerationmethod}, lacking the skip connections found in U-Net structures, makes traditional feature map caching methods less effective. We propose $\ours$, a novel training-free inference acceleration framework that jointly optimizes hierarchical latent caching, adaptive importance-guided quantization, and structural redundancy-aware pruning.
\vspace{-1.5mm}
\subsection{Joint Optimization of HLC and AIGQ}
\vspace{-1.5mm}

To enhance the efficiency of Diffusion Transformers (DiTs) in video generation, we propose a joint optimization framework that integrates \textit{Hierarchical Latent Caching} with \textit{Adaptive Importance-Guided Quantization}, as shown in~\cref{fig:overview}. This framework stems from a novel observation: within the iterative denoising process, the relative importance of different layers and timesteps fluctuates dynamically, depending on the evolution of latent representations and cross-step feature interactions. Crucially, we find that certain layers play a pivotal role in refining spatial structure, while others primarily contribute to temporal smoothness. Similarly, the degree of information retention across timesteps varies non-uniformly, creating opportunities for selectively caching features and adjusting quantization granularity based on their functional relevance.

\noindent\textbf{Hierarchical Latent Caching.$\quad$}
Unlike conventional caching mechanisms relying on static cache intervals, inspired by~\cite{chen2024deltadittrainingfreeaccelerationmethod,adacache,liu2025fasterdiffusiontemporalattention,li2024fasterdiffusionrethinkingrole}, our approach leverages an adaptive refresh strategy that dynamically determines where recomputation is necessary. Given that DiTs lack the explicit skip connections found in U-Net architectures, we model cache decisions using an importance-aware metric that considers inter-step feature variations. Specifically, at timestep $t$, we compute a timestep-wise feature divergence score $\mathcal{D}_t^{(l)}$:
\begin{equation}
\mathcal{D}_t^{(l)} = \frac{\| p_t^{(l)} - p_{t-k}^{(l)} \|_1}{k}  \cdot \| \nabla_t m_t^{(l)} \|,
\label{eq1}
\end{equation}
where $p_t^{(l)}$ represents the activation at layer $l$ and timestep $t$, $k$ is the last cached step, and $\nabla_t m_t^{(l)}$ denotes the inter-frame gradient of the feature map, capturing the rate of change in motion across consecutive timesteps. 

Based on the feature divergence score, we establish a \textit{cache-refresh decision function}:
\begin{equation}
\tau_t^{(l)} = 
\begin{cases}
\tau_{\max}, & \text{if } \mathcal{D}_t^{(l)} < \delta_1, \\
\tau_{\text{mid}}, & \text{if } \delta_1 \leq \mathcal{D}_t^{(l)} < \delta_2, \\
\tau_{\min}, & \text{if } \mathcal{D}_t^{(l)} \geq \delta_2,
\end{cases}
\label{eq2}
\end{equation}
where $\tau_t^{(l)}$ determines the number of steps before recomputation, adapting caching frequency to content variations.

\begin{figure}[t]
\centering
\includegraphics[width=1\linewidth]{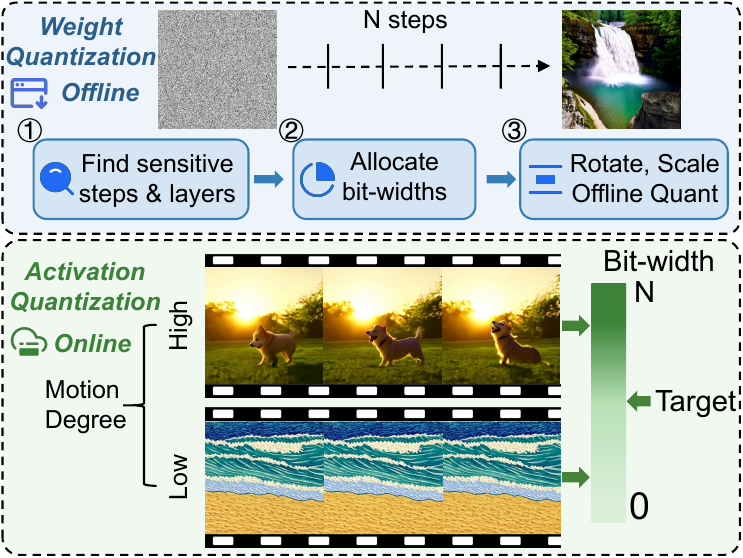}
\vspace{-7mm}
\caption{AIGQ: adaptive importance-guided quantization.}
\label{fig:aigq}
\vspace{-6mm}
\end{figure}

\noindent\textbf{Adaptive Importance-Guided Quantization.$\quad$}
While caching effectively reduces redundant computations, quantization offers an orthogonal opportunity to optimize inference efficiency. However, naive low-bit quantization risks degrading spatial and temporal coherence, particularly in critical layers. To address this, we introduce an importance-driven mixed-precision quantization scheme that dynamically assigns bit-widths to both weights and activations based on their perceptual relevance, as shown in~\cref{fig:aigq}.

\textit{(1) Weight Quantization}: Instead of applying a uniform bit-width across all layers, inspired by~\cite{shang2023pbllmpartiallybinarizedlarge,zhao2024viditq,
jo2024mixturescalesmemoryefficienttokenadaptive}, we allocate a quantization budget $B_{\text{total}}$ by first estimating layer sensitivity. Specifically, we evaluate each layer based on its numerical error, perceptual distortion, and temporal dynamics. Layers that contribute significantly to fine-grained texture reconstruction or motion continuity are assigned higher precision, whereas those with minimal impact on perceptual quality are allocated lower bit-widths.

To further mitigate quantization-induced artifacts, we incorporate a channel-balancing mechanism that combines scaling and rotation. The scaling-based approach corrects static imbalances originating from pretrained scale shift tables, while the rotation-based method addresses dynamic variations caused by timestep embeddings. By first applying scaling to stabilize the initial activation distribution, followed by a lightweight rotation transformation, we ensure a more uniform data distribution across channels, reducing extreme outliers that could degrade quantization performance. Given the total bit-width budget, we iteratively allocate precision levels while satisfying:
\begin{equation}
\sum_{l} B(l) \leq B_{\text{total}},
\end{equation}
where bit-widths are assigned to layers that exhibit higher sensitivity to quantization. This adaptive allocation ensures that critical layers retain sufficient precision while computational efficiency is maximized, enabling effective compression with minimal impact on generative quality.

\textit{(2) Activation Quantization}: Beyond optimizing weight precision, we extend our quantization strategy to activation, where bit-widths are dynamically modulated based on timestep-level redundancy. Inspired by~\cite{dong2024ditas,chen2024deltadittrainingfreeaccelerationmethod,adacache}, We observe that not all timesteps contribute equally to the final output quality. In early stages or during redundant intermediate steps, feature representations often exhibit high similarity, suggesting that lower precision is sufficient without compromising perceptual fidelity. Conversely, during critical transitions—such as the emergence of fine details or significant structural changes—higher precision is essential to capture the complexity of the evolving content. 

Based on the observation, we propose a novel timestep-wise content-adaptive bit allocation function that tailors activation bit-widths to the specific demands of each step, thereby optimizing both computational efficiency and output quality. Formally, our allocation function is defined as:
\begin{equation}
\text{bit-width}(t) =
\begin{cases}
Bit_{\max}, & \text{if } \mathcal{D}_t < \theta_1, \\
Bit_{\text{mid}}, & \text{if } \theta_1 \leq \mathcal{D}_t < \theta_2, \\
Bit_{\min}, & \text{if } \mathcal{D}_t \geq \theta_2, \\
\end{cases}
\end{equation}
where $\mathcal{D}_t$ represents a timestep-specific redundancy metric (\textit{e.g.} distance between consecutive feature maps), and $\theta_1$ and $\theta_2$ are empirically determined thresholds that delineate low, medium, and high redundancy regimes. Here, $Bit_{\max}$, $Bit_{\text{mid}}$, and $Bit_{\min}$ denote the maximum, intermediate, and minimum bit-widths, respectively.

\begin{figure*}[t]
    \centering
    \begin{minipage}{0.48\linewidth}
        \centering        
        \vspace{-6mm}
        \includegraphics[width=\linewidth]{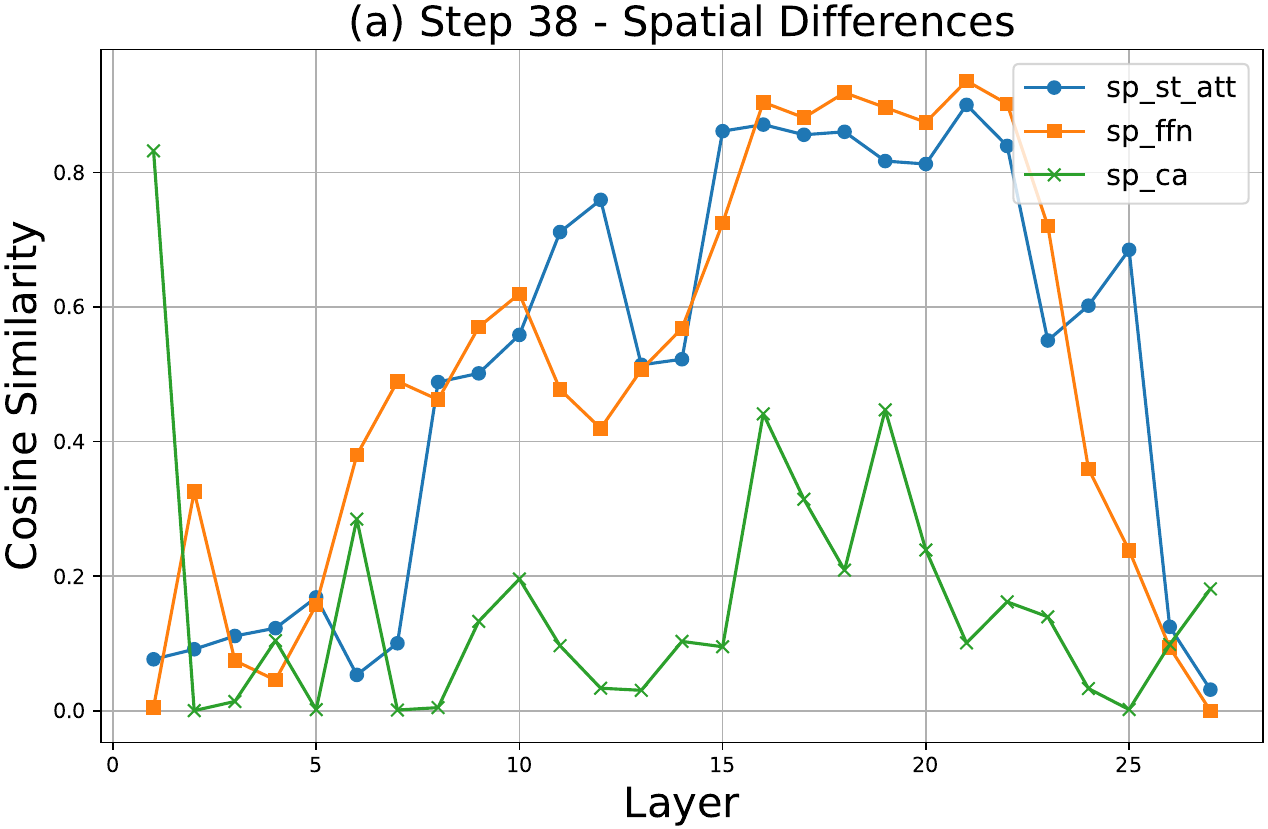}
        \vspace{-6mm}        
    \end{minipage}
    \hfill
    \begin{minipage}{0.48\linewidth}
        \centering
        \vspace{-6mm}
        \includegraphics[width=\linewidth]{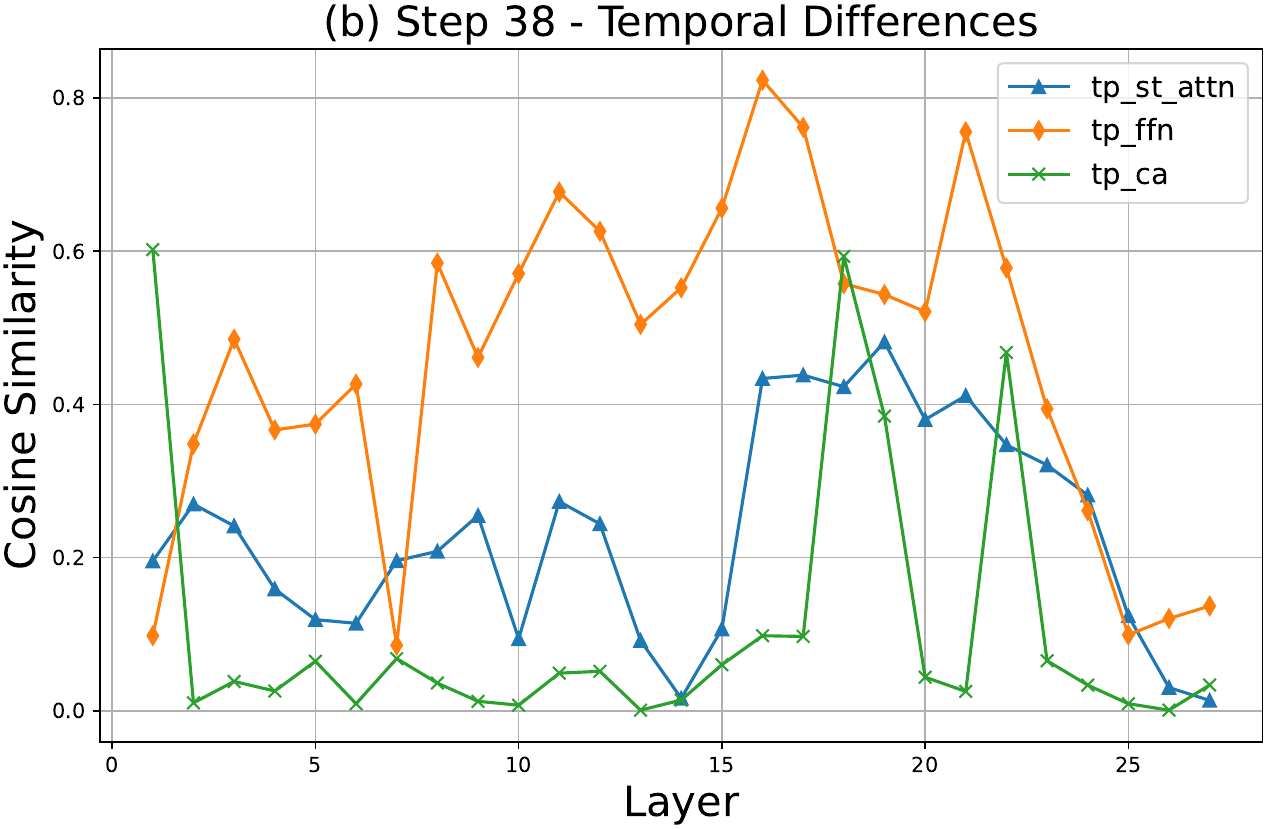}
        \vspace{-6mm}
    \end{minipage}
    \vspace{-1.5mm}
\caption{Spatial and temporal differences across adjacent layers for spatial-temporal attention, cross-attention, and feed-forward network.}

    \vspace{-5mm}
    \label{fig:srp}
\end{figure*}

The intuition behind this design is straightforward yet powerful: steps with high feature redundancy (i.e., $\mathcal{D}_t \geq \theta_2$) can tolerate aggressive quantization to $Bit_{\min}$, as the information loss is minimal and does not degrade the generative process. In video generation, consecutive frames with subtle changes—like a static background—require less precision in activations, allowing us to allocate fewer bits without sacrificing visual coherence. In contrast, steps with low redundancy (i.e., $\mathcal{D}_t < \theta_1$), such as those involving abrupt scene transitions or the refinement of intricate textures, demand $Bit_{\max}$ to preserve the fidelity of complex features. The intermediate range ($Bit_{\text{mid}}$) serves as a balanced compromise for timesteps with moderate complexity, ensuring a smooth trade-off between efficiency and quality.

This adaptive strategy reduces memory footprint and computational overhead, and aligns quantization decisions with the intrinsic dynamics of the generative model. By adjusting activation precision dynamically, we avoid pitfalls of uniform quantization, which over-allocates resources to redundant steps or under-allocates to critical ones. 

\noindent\textbf{Unified Optimization for Efficient Video Generation.$\quad$}  
By integrating \textit{Hierarchical Latent Caching} (HLC) with \textit{Adaptive Importance-Guided Quantization} (AIGQ), we construct a self-adaptive compute allocation strategy that minimizes unnecessary computation while preserving video generation quality. HLC uses $\mathcal{D}_t^{(l)}$ in~\cref{eq1} to assess feature divergence at timestep $t$, guiding adaptive caching. AIGQ leverages a related timestep-wise metric $\mathcal{D}_t$, derived from $\mathcal{D}_t^{(l)}$ via feature similarity, to dynamically adjust activation bit-widths. For smaller skips (low $\tau_t^{(l)}$ in~\cref{eq2}), we use smaller bit-widths to exploit redundancy, while for larger skips (high $\tau_t^{(l)}$ in~\cref{eq2}), we apply smaller bit-widths to enhance precision in post-skip steps, ensuring video quality. This cohesive timestep-wise strategy optimizes computation and precision.

\begin{table*}[t]
\footnotesize
\centering
\renewcommand{\arraystretch}{0.98} 
\resizebox{\textwidth}{!}{%
\begin{tabular}{l c c c c c c c c c c} 
\toprule
{Method} & {\makecell{Bit-width \\ (W/A)}} & {\makecell{Motion \\ Smooth.}} & {\makecell{BG. \\ Consist.}} & {\makecell{Subject \\ Consist.}} & {\makecell{Aesthetic \\ Quality}} & {\makecell{Imaging \\ Quality}} & {\makecell{Dynamic \\ Degree}}& {\makecell{Scene \\ Consist.}} & {\makecell{Overall \\ Consist.}} \\
\midrule \midrule
Open-Sora~\cite{opensora}    & 16/16 & 98.42 & 96.44 & 95.20 & 60.07 & 59.66 & 33.33&41.72 & 26.89 \\
\midrule
Q-diffusion~\cite{li2023qdiffusion}  & 8/8  & 96.54 & 94.47 & 92.52 & 58.00 & 56.57 & 38.88 & 38.57 & 26.33 \\
Q-DiT~\cite{chen2024QDiT}        & 8/8  & 95.72 & 95.01 & 91.68 & 58.68 & 56.54 & 38.88 & 34.06 & 26.77 \\
PTQ4DiT~\cite{NEURIPS2024_72d32f4f}      & 8/8  & 98.02 & 96.33 & 96.23 & 58.40 & 53.29 & 37.50 & 36.36 & 25.98 \\
SmoothQuant~\cite{xiao2023smoothquant}  & 8/8  & 98.09 & 94.47 & 92.49 & 58.79 & 58.29 & 38.88 & 38.61 & 26.33 \\
Quarot~\cite{ashkboos2024quarot}       & 8/8  & 97.09 & 95.34 & 90.00 & 55.96 & 56.34 & 37.50 & 37.55 & 26.09 \\
ViDiT-Q~\cite{zhao2024viditq}      & 8/8  & 98.28 & 96.15 & 95.16 & 59.89 & 59.47 & 34.72 & 40.26 & 26.74 \\
\rowcolor{orange} $\ours$ & 8/8 & 98.52 & 96.12 & 94.62 & 58.57 & 55.94 & 31.94& 36.92 & 26.97 \\
\midrule

Q-DiT~\cite{chen2024QDiT}     & 4/8  & 99.88 & 97.33 & 96.50 & 31.14 & 21.83 & 2.77 & 0.00  & 5.11  \\
PTQ4DiT~\cite{NEURIPS2024_72d32f4f}      & 4/8  & 94.62 & 98.50 & 98.69 & 32.76 & 35.57 & 5.56 & 3.75  & 11.76 \\
SmoothQuant~\cite{xiao2023smoothquant}  & 4/8  & 96.69 & 94.66 & 97.85 & 46.67 & 44.01 & 12.50 & 27.82 & 18.72 \\
Quarot~\cite{ashkboos2024quarot}       & 4/8  & 94.63 & 94.55 & 99.70 & 46.04 & 41.46 & 37.50 & 29.94 & 18.91 \\
ViDiT-Q~\cite{zhao2024viditq}      & 4/8  & 97.82 & 95.54 & 93.55 & 58.23 & 57.21 & 33.33 &38.12 & 26.61 \\
\rowcolor{orange} $\ours$ & 4/6 & 98.57 & 96.34 & 94.56 & 58.63 & 55.94 & 34.72 & 39.39 & 26.77 \\
\bottomrule
\end{tabular}%
}
\vspace{-3mm}
\caption{Performance comparison of various methods on VBench~\cite{huang2023vbench,huang2024vbench++}. The bit-width ``16" refers to FP16 without quantization, while $\ours$-4/6 represents the version with adaptive importance-guided quantization. Due to failure to generate readable content, Q-diffusion for W4A8 is omitted. Notably, $\ours$-4/6 shows negligible loss in quality metrics compared to the baseline Open-Sora. 
}
\label{tab:comparison}
\vspace{-6mm}
\end{table*}

\vspace{-1.5mm}
\subsection{Structural Redundancy-Aware Pruning}
\vspace{-1.5mm}
To further enhance the efficiency of DiTs, we propose a novel \textit{Structural Redundancy-Aware Pruning (SRAP)} mechanism. In~\cref{fig:srp}, SRAP adaptively prunes layers within a single timestep based on their internal feature similarity. Unlike conventional static layer pruning strategies that predefine a fixed subset of layers to be pruned, SRAP dynamically determines layer importance \textit{at runtime} by analyzing the intrinsic similarity of intermediate representations. This allows us to minimize redundant computations.

\noindent\textbf{Motivation: Layer-Wise Redundancy in DiTs.$\quad$}
Studies in vision transformers show that early layers capture coarse-grained details, while deeper layers refine high-frequency components~\cite{ghiasi2022vision,zhang2022nested}. However, in DiTs, Feature evolution is highly iterative across timesteps. We observe that \textit{certain layers within a single timestep exhibit significant representational overlap}. This phenomenon suggests that some computations can be pruned without information  loss. By quantifying this redundancy, we introduce a data-driven mechanism to selectively prune layers at inference time.

\noindent\textbf{Quantifying Redundancy with Cosine Similarity.$\quad$}
To determine which layers can be pruned, inspired by~\cite{dumitru2024changeconstantdynamicllm,men2024shortgptlayerslargelanguage}, we compute a layer-wise cosine similarity score between consecutive layers at timestep $t$~(\cref{fig:srp}):
\begin{equation}
S_t^{(l, l+1)} = \frac{\langle p_t^{(l)}, p_t^{(l+1)} \rangle}{\| p_t^{(l)} \| \| p_t^{(l+1)} \|},
\end{equation}
where $p_t^{(l)}$ and $p_t^{(l+1)}$ represent the feature representations at layers $l$ and $l+1$ within timestep $t$. The closer this score is to 1, the higher the redundancy between layers.

Based on the cosine similarity measure, we define a \textit{layer pruning probability function}: 
\begin{equation}
P_{\text{prune}}^{(l)} = 
\begin{cases}
1, & \text{if } S_t^{(l, l+1)} > \tau_{\text{high}}, \\
P_{\text{base}}, & \text{if } \tau_{\text{low}} \leq S_t^{(l, l+1)} \leq \tau_{\text{high}}, \\
0, & \text{if } S_t^{(l, l+1)} < \tau_{\text{low}},
\end{cases}
\end{equation}
where $\tau_{\text{high}}$ and $\tau_{\text{low}}$ define redundancy thresholds, and $P_{\text{base}}$ is a baseline probability allowing occasional pruning when similarity is moderate. If two layers exhibit strong redundancy ($S_t^{(l, l+1)} > \tau_{\text{high}}$), we completely bypass computations for $p_t^{(l+1)}$, reducing the inference cost.

\noindent\textbf{Adaptive Layer Pruning Strategy.$\quad$}
Rather than applying uniform layer pruning across all timesteps, we introduce an \textit{adaptive pruning mechanism} that considers temporal feature dynamics. Specifically, we track the cumulative feature variation across previous timesteps:
\begin{equation}
\mathcal{V}_t = \sum_{i=0}^{k} \| p_t - p_{t-i} \|_1.
\end{equation}
If $\mathcal{V}_t$ is below a threshold $\delta_{\text{low}}$, indicating that the diffusion process is carefully refining rather than radically transforming content, we increase the layer pruning probability across all redundant layers. Conversely, if $\mathcal{V}_t$ exceeds $\delta_{\text{high}}$, meaning that drastic changes are actively occurring, we reduce layer pruning to maintain information flow.

\noindent\textbf{Joint Optimization with Quantization and Caching.$\quad$}
The integration of Structural Redundancy-Aware Pruning (SRAP) with our existing quantization and caching strategies creates a \textit{three-tier compute optimization framework}:
(1) Hierarchical Latent Caching eliminates redundant computations across timesteps.
(2) Adaptive Importance-Guided Quantization dynamically reduces numerical precision based on feature sensitivity.
(3) Structural Redundancy-Aware Pruning selectively prunes layers within a timestep to prevent unnecessary overhead. 
 
By holistically optimizing compute allocation across layers and timesteps, our method significantly accelerates DiT inference while preserving generative fidelity. This marks a substantial step forward in efficient video diffusion model and deployment for real-world applications.

\begin{table}[t]

\renewcommand{\arraystretch}{0.98} 
        \hfill
        \centering
        \setlength{\tabcolsep}{1mm}
        \resizebox{1\linewidth}{!}{
            \begin{tabular}{lccccc}
            \toprule[1pt]
            \multirow{2}{*}{{Method}} & {Bit-width} & \multirow{2}{*}{{CLIPSIM}} &  {CLIP-} & VQA- & VQA-  \\
             & (W/A) & & Temp & \small{Aesthetic} & \small{Technical}  \\
            \midrule \midrule
            Open-Sora~\cite{opensora} & 16/16 & 0.1842 & 0.9983 & 62.58 & 50.18  \\
            \cmidrule(lr){1-6}
             Q-DiT~\cite{chen2024QDiT} & 8/8 & 0.1833 & 0.9972 & 60.24 & 34.78  \\
             PTQ4DiT~\cite{NEURIPS2024_72d32f4f} & 8/8 & 0.1882 & 0.9986 & 53.85 & 53.03  \\
             SmoothQuant~\cite{xiao2023smoothquant} & 8/8 & 0.2000 & 0.9981 & 59.01 & 51.24   \\
             Quarot~\cite{ashkboos2024quarot} & 8/8 & 0.1990 & 0.9971 & 57.97 & 51.99   \\
              
             ViDiT-Q~\cite{zhao2024viditq} & 8/8 & 0.1999 & 0.9986 & 59.91 & 54.34   \\
             \rowcolor{orange}$\ours$ & 8/8 & 0.1925 & 0.9989 & 60.19 & 52.39 \\
            
            \cmidrule(lr){1-6}
              Q-DiT~\cite{chen2024QDiT} & 4/8 & 0.1729 & 0.9828 & 0.01 & 0.02 \\
              PTQ4DiT~\cite{NEURIPS2024_72d32f4f} & 4/8 & 0.1778 & 0.9968 & 2.18 & 0.32 \\
              SmoothQuant~\cite{xiao2023smoothquant} & 4/8 & 0.1878 & 0.9978 & 90.77 & 22.72\\
              Quarot~\cite{ashkboos2024quarot} & 4/8 & 0.1863 & 0.9960 & 46.75 & 32.95 \\
               
              ViDiT-Q~\cite{zhao2024viditq} & 4/8 & 0.1854 & 0.9984 & 59.84 & 49.11   \\
              \rowcolor{orange}$\ours$ & 4/6 & 0.1904 & 0.9981 & 59.92 & 49.14 \\
            \bottomrule[1pt]
            \vspace{-10pt}
            \end{tabular}
            }
            \vspace{-3.5mm}
            \caption{Performance comparison of various methods on CLIP and Dover. The bit-width ``16" refers to FP16 without quantization, while $\ours$-4/6 represents the version with adaptive importance-guided quantization.}
            \label{tab:quality_comparison}
            \vspace{-7mm}
\end{table}

\begin{table*}[t]
\centering
\resizebox{\textwidth}{!}{
\begin{tabular}{c@{\hspace{10mm}}c@{\hspace{10mm}}cccccccc}
\toprule[1pt]
\multicolumn{3}{c}{{Methods}} & \multirow{2}{*}{{Motion Smooth.}} & \multirow{2}{*}{{BG. Consist.}} & \multirow{2}{*}{{Subject Consist.}} & \multirow{2}{*}{{Aesthetic Quality}} & \multirow{2}{*}{{Imaging Quality}} & \multirow{2}{*}{{Speedup}} \\
\cmidrule(lr){1-3} 
{HLC} & {AIGQ} & {SRAP} & & & & & & \\
\midrule \midrule
- & - & - & 99.29 & 98.10 & 97.74 & 63.09 & 59.37 & 1.00$\times$ \\
\checkmark & - & - & 99.21 & 97.59 & 97.65 & 62.09 & 58.28 & 4.12$\times$ \\
\checkmark & \checkmark & - & 99.16 & 97.62 & 97.62 & 61.61 & 55.68 & 6.33$\times$ \\
\checkmark & \checkmark & \checkmark & 98.91 & 96.19 & 97.29 & 61.39 & 55.64 & 6.72$\times$ \\
\bottomrule[1pt]
\end{tabular}}
\vspace{-3mm}
\caption{Ablation studies. Evaluation on motion smoothness, background consistency, subject consistency, aesthetic quality, imaging quality, and speedup demonstrates the proposed HLC, AIGQ, and SRAP achieve significant speedup with minimal performance degradation.}
\label{tab:ablation}
\vspace{-6mm}
\end{table*}

\vspace{-2mm}
\section{Experiment}
\vspace{-2mm}

\subsection{Experiment Settings}
\vspace{-2mm}
We evaluate the effectiveness of $\ours$ on Open-Sora1.2~\cite{opensora}, exploring different bit-width configurations and acceleration strategies. The videos are generated with 100 timesteps. More comprehensive discussion of implementation details are provided in supplementary materials.

\noindent\textbf{Quantization Scheme.$\quad$} We employ a uniform min-max quantization with per-channel weight and dynamic per-layer activation quantization. The activation quantization parameters are computed online with minimal computational overhead, ensuring adaptability across varying feature distributions. Our mixed-precision weight quantization is determined offline using a small calibration dataset, balancing numerical efficiency with generation quality.

\noindent\textbf{Evaluation Settings.$\quad$}
We evaluate the performance of $\ours$ using the VBench benchmark suite~\cite{huang2023vbench,huang2024vbench++}, which provides a comprehensive set of evaluation metrics. In alignment with prior works~\cite{ren2024consisti2v,zhao2024viditq} , we select 8 key evaluation dimensions from VBench to ensure a thorough assessment. Furthermore, we adopt CLIP used in~\cite{liu2023evalcrafter} and Dover~\cite{wu2023dover} and  benchmarks, chosen based on their relevance to our experimental objectives. Specifically, we use CLIPSIM and CLIP-Temp to measure the alignment between text and video, as well as to assess temporal semantic consistency. Additionally, we utilize DOVER for video quality assessment, which evaluates the generation quality from both aesthetic perspectives and technical metrics.

\noindent\textbf{Hardware Implementation.$\quad$}
To efficiently implement $\ours$ in hardware, we developed optimized GEMM CUDA kernels that handle both quantization and caching mechanisms, resulting in better resource utilization and improved inference speed. Inspired by \cite{lin2024qserve,xiao2023smoothquant,zhao2024viditq}, we absorb the scaling-based channel balancing factors into the preceding layers offline to enhance computational efficiency. Additionally, we apply kernel fusion, which combines the quantization process with rotation transformations, while leveraging intermediate feature caching. The optimized CUDA kernels effectively reduce the computational cost of $\ours$, achieving a \textbf{6.72$\times$} speedup on a single NVIDIA A800-80GB GPU with CUDA 12.1.

\vspace{-2mm}
\subsection{Main Results}
\vspace{-2mm}
Tables~\ref{tab:comparison} and~\ref{tab:quality_comparison}, demonstrate the significant improvements achieved by $\ours$, over other SOTA methods.

\noindent\textbf{VBench Quality Comparison.$\quad$}
First, we analyze the quality of the generated video frames using VBench~\cite{huang2023vbench, huang2024vbench++}~(\cref{tab:comparison}). In terms of bit-width, $\ours$ operates with bit-widths of 8/8 and 4/6 for weights and activations, comparable to other methods like Q-diffusion~\cite{li2023qdiffusion}, Q-DiT~\cite{chen2024QDiT}, PTQ4DiT~\cite{NEURIPS2024_72d32f4f}, and SmoothQuant~\cite{xiao2023smoothquant}.
Specifically, for the 8/8 bit-width setting, $\ours$ achieves strong performance, with only minor reductions compared to the baseline Open-Sora~\cite{opensora} model. In the 4/8 bit-width setup, Q-DiT and PTQ4DiT struggle to maintain content quality. In the more challenging 4/6 bit-width configuration, $\ours$ still outperforms other methods with 4/8 bit-width , showing the model's robustness across different precision settings. 

\noindent\textbf{CLIP and Dover Quality Comparison.$\quad$}
Table~\ref{tab:quality_comparison} demonstrates the generated outputs using CLIP and Dover metrics, including CLIPSIM, CLIP-Temp, and two VQA tasks: Aesthetic and Technical. $\ours$ achieves a higher aesthetic and technical score in both the 8/8 and 4/6 bit-width configurations, demonstrating its ability to generate high-quality video content even with reduced bit-widths. For more detailed comparisons across additional prompts, please refer to the supplementary file.

\vspace{-1.5mm}
\subsection{Ablation Studies}
\vspace{-1.5mm}
We present ablation studies in Tab.~\ref{tab:ablation} to evaluate the contribution of different components in our proposed framework. We use the Open-Sora~\cite{opensora} prompt sets for video generation and select five representative evaluation metrics from VBench~\cite{huang2023vbench,huang2024vbench++} for performance assessment. We begin by evaluating the baseline configuration, where no enhancements are applied, providing a reference point for assessing the impact of each technique.

\noindent\textbf{Evaluation of \textit{HLC}.$\quad$}
As shown in ~\cref{tab:ablation}, enabling \textit{HLC} improves efficiency by reducing redundant timesteps, resulting in a notable speedup of 4.12$\times$. Quality metrics show minimal degradation, with only slight reductions observed in comparison to the baseline.

\noindent\textbf{Evaluation of \textit{AIGQ}.$\quad$}
Next, we incorporate \textit{AIGQ} alongside HLC. The dynamic allocation of precision to weights and activations based on their importance further refines the model's performance, leading to a higher speedup of 6.33$\times$ with negligible visual degradation.

\noindent\textbf{Evaluation of \textit{SRAP} and Full Model.$\quad$}
We evaluate the full model, where SRAP selectively prunes redundant layers to contribute to further acceleration. The complete implementation of $\ours$ achieves a speedup of 6.72$\times$. Despite slight quality loss, this configuration provides the best trade-off between efficiency and generation quality.

\begin{table}[t]
\footnotesize
\centering
\renewcommand{\arraystretch}{0.9} 
\setlength{\tabcolsep}{1.5mm} 
    \resizebox{\linewidth}{!}{
        \begin{tabular}{lccc}
            \toprule[1pt]
            {Method} & {Bitwidth (W/A)} & ~~{Cache}~~ & ~~{Speedup}~~ \\
            \midrule \midrule
            Open-Sora~\cite{opensora} & 16/16 & -  & 1.00 $\times$  \\ \midrule
            T-Gate~\cite{tgate} & 16/16 & \checkmark & 1.10 $\times$ \\
            PAB~\cite{zhao2024real} & 16/16 & \checkmark & 1.34 $\times$ \\
            ViDiT-Q~\cite{zhao2024viditq}  & 8/8 & - & 1.71 $\times$ \\
            AdaCache-slow~\cite{adacache} & 16/16 & \checkmark & 1.46 $\times$ \\
            AdaCache-fast~\cite{adacache} & 16/16 & \checkmark & 2.24 $\times$ \\           
            \rowcolor{orange} $\ours$ &  4/6& \checkmark & 6.72 $\times$ \\
            \bottomrule[1pt]
        \end{tabular}
    }

   \vspace{-3mm}
    \caption{Speedup performance comparison of various methods, including the impact of bitwidth and cache on their performance.}
    \label{tab:acceleration_comparison}
    \vspace{-5mm}
\end{table}

\vspace{-1.5mm}
\subsection{Speedup Performance}
\vspace{-1mm}

As shown in Tab.~\ref{tab:acceleration_comparison}, T-Gate~\cite{tgate}, PAB~\cite{zhao2024real}, ViDiT-Q~\cite{zhao2024viditq}, and AdaCache~\cite{adacache} provide more efficient solutions compared to the baseline Open-Sora~\cite{opensora}. The speedup for these methods ranges from 1.10$\times$ (for T-Gate) to 2.24$\times$ (for AdaCache-fast), while $\ours$ achieves a remarkable speedup of 6.72$\times$, significantly surpassing all other methods while maintaining high generation quality. This substantial improvement is attributed to our low bit-width quantization and sophisticated caching strategies.

\noindent\textbf{CUDA Acceleration. $\quad$} A key factor in $\ours$’s superior performance lies in its ability to balance low bit-width quantization with caching, incorporating kernel fusion techniques in our CUDA implementation for enhanced computational efficiency. These kernels integrate quantization with rotation transformations and intermediate feature caching. $\ours$ maximizes GPU resource utilization, resulting in both faster computation and lower latency, making it particularly well-suited for efficient video generation.

\vspace{-2mm}
\section{Conclusion}
\vspace{-1.5mm}
We propose a joint optimization framework that integrates hierarchical latent caching, adaptive importance-guided quantization, and structural redundancy-aware pruning to accelerate Diffusion Transformers (DiTs) for video generation. By adaptively reusing cached features, adjusting bit-widths based on content sensitivity, and pruning redundant layers, our method achieves efficient inference while maintaining high generation quality. Our approach achieves a \textbf{6.72$\times$} speedup on Open-Sora with minimal degradation in generation quality. We believe $\ours$ provides a scalable and efficient solution for accelerating DiTs, making high-fidelity video generation more accessible for real-world and resource-constrained applications.

{
    \small
    \bibliographystyle{ieeenat_fullname}
    \bibliography{main}
}


\end{document}